\documentclass[dvipsnames,format=sigconf,anonymous=false,review=false]{acmart}\usepackage{graphicx} 
\usepackage{todonotes}

\title[Instance Selection for Dynamic Algorithm Configuration with RL]{Instance Selection for Dynamic Algorithm Configuration with Reinforcement Learning: Improving Generalization}

\author[Benjamins]{Carolin Benjamins}
\affiliation{%
 \institution{Leibniz University Hannover}
 \country{Germany}
 }
\author[Cenikj]{Gjorgjina Cenikj}
\affiliation{%
 \institution{Jožef Stefan International Postgraduate School\\Jožef Stefan Institute}
 \country{Slovenia}
 }
\author[Nikolikj]{Ana Nikolikj}
\affiliation{%
 \institution{Jožef Stefan International Postgraduate School\\Jožef Stefan Institute}
 \country{Slovenia}
 }
\author[Mohan]{Aditya Mohan}
\affiliation{%
 \institution{Leibniz University Hannover}
 \country{Germany}
 }
\author[Eftimov]{Tome Eftimov}
\affiliation{%
 \institution{Jožef Stefan Institute}
 \country{Slovenia}
 }
\author[Lindauer]{Marius Lindauer}
\affiliation{%
 \institution{Leibniz University Hannover\\ L3S Research Center}
 \country{Germany}
 }

\usepackage[noabbrev,capitalize]{cleveref}
\usepackage[nolist]{acronym}
\usepackage{bm}
\usepackage{comment}
\usepackage{amsmath}
\usepackage{mathtools}  
\usepackage{amsfonts} 
\usepackage{paralist}
\usepackage{enumitem}  
\usepackage{placeins}

\newcommand{\algo}{A}
\newcommand{\insts}{\mathcal{I}}
\newcommand{\trainset}{\insts_{train}}
\newcommand{\testset}{\insts_{test}}
\newcommand{\selset}{\insts_{selected}}
\DeclareMathOperator*{\argmax}{arg\,max}        

\begin{acronym}
\acro{DAC}{Dynamic Algorithm Configuration}
\acro{RL}{Reinforcement Learning}
\acro{MDP}{Markov Decision Process}
\acro{PPO}{Proximal Policy Optimization}
\end{acronym}

\copyrightyear{2024}
\acmYear{2024}
\setcopyright{rightsretained}
\acmConference[GECCO '24 Companion]{Genetic and Evolutionary Computation Conference}{July 14--18, 2024}{Melbourne, VIC, Australia}
\acmBooktitle{Genetic and Evolutionary Computation Conference (GECCO '24 Companion), July 14--18, 2024, Melbourne, VIC, Australia}
\acmDOI{10.1145/3638530.3654291}
\acmISBN{979-8-4007-0495-6/24/07}

\begin{document}
\begin{abstract}
Dynamic Algorithm Configuration (DAC) addresses the challenge of dynamically setting hyperparameters of an algorithm for a diverse set of instances rather than focusing solely on individual tasks.
Agents trained with Deep Reinforcement Learning (RL) offer a pathway to solve such settings.
However, the limited generalization performance of these agents has significantly hindered the application in DAC.
Our hypothesis is that a potential bias in the training instances limits generalization capabilities.
We take a step towards mitigating this by selecting a representative subset of training instances to overcome overrepresentation and then retraining the agent on this subset to improve its generalization performance. 
For constructing the meta-features for the subset selection, we particularly account for the dynamic nature of the RL agent by computing time series features on trajectories of actions and rewards generated by the agent's interaction with the environment.
Through empirical evaluations on the Sigmoid and CMA-ES benchmarks from the standard benchmark library for DAC, called DACBench, we discuss the potentials of our selection technique compared to training on the entire instance set.
Our results highlight the efficacy of instance selection in refining DAC policies for diverse instance spaces.
\end{abstract}


\maketitle

\section{Introduction}

\ac{DAC} offers an automated solution to the task of setting algorithm hyperparameters dynamically, by determining well-performing hyperparameter schedules or policies.
One way to learn such policies is through \acf{RL}~\citep{biedenkapp-ecai20a,adriaensen-jair22a}.
While conceptually appealing, \ac{RL} algorithms have the notorious tendency to significantly overfit their training environments~\cite{zhang-arxiv18b,justesen-neurips18a,kirk-jair23a}.
As a consequence, RL methods for \ac{DAC} suffer from a lack of generalization to instances not seen during training, thereby limiting their applicability.


We take a step towards improving the generalization performance of RL policies on new test instances by subselecting representative training instances using SELECTOR~\cite{cenikj-gecco22a}.
To capture the dynamic nature of RL, we use trajectory-based representations generated by the RL algorithm after training on the full instance~set.

\begin{figure*}[t]
    \centering
    \includegraphics[width=0.9\linewidth]{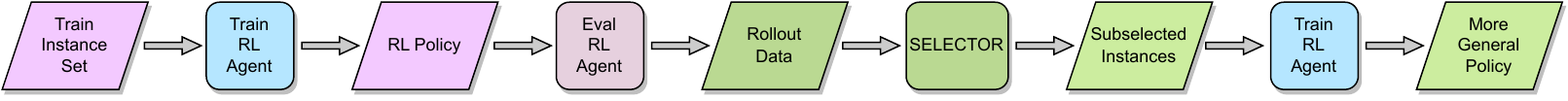}
    \caption{Our proposed flow of subselecting representative instances with SELECTOR for DAC with RL
    }
    \label{fig:instance_dac_method}
\end{figure*}



Concretely, we make the following contributions: i) For \ac{DAC} with \ac{RL}, we present a principled framework to select representative instances to train on to improve generalization to the instance space; ii) we propose a new domain-agnostic approach for generating instance meta-features that encode the dynamics of the DAC problem; iii) we demonstrate superior performance training on the subselected instance set; iv) we analyze the selected instances; and v) we provide an insight on how to use the framework SELECTOR.

\textbf{Reproducibility:} Code and data is available here: \url{https://github.com/automl/instance-dac}.

\section{Method}
\label{sec:method}

The goal of our study is to improve the generalization of an \ac{RL} agent in \acf{DAC}, as measured by the performance of a policy on a test set of target instances.
\cref{fig:instance_dac_method} shows the outline of our method.
Overall, we use SELECTOR to sample a subset of representative training instances, to which we then allocate more training resources.
We fix the total number of times the \ac{RL} agent interacts with the environment before the subselection and after the subselection to be the same.
This means for the same training budget, we train on fewer but more representative instances after subselection.

To enable this workflow, we start by training an \ac{RL} agent on the train instance set $\trainset$.
A key element is using meta-features based on the data from the trajectory generated by the \ac{RL} agent as it interacts with the algorithm.
We do this by evaluating the trained agent on the train instance set and producing rollout trajectories, specifically the actions taken by the agent and the reward received for each action.
This data encodes the agent's behavior for each training instance.
We feed these meta-feature data for all training instances to SELECTOR, which subselects instances from the train instance set to form the reduced, subselected instance set $\selset \subseteq \trainset$.
Intuitively, these instances capture the essential aspects of the dynamics observed by the agent during training and should, therefore, enable better generalization. 
We finally train the \ac{RL} agent again on the subselected instance set to obtain the final policy, which we can subsequently evaluate on the held-out test set of instances.

\paragraph{Meta-Feature Representations}
SELECTOR requires the representation of data instances (in our case, episodes from training the RL agent) to be numerical features. 
In prior work, instance meta-features were obtained via a manual approach~\cite{bischl-aij16a}, possibly not always reflecting the agent's interaction with the environment.
Our approach, however, uses features from the data generated by the agent as it interacts, thus capturing the agent's dynamic behavior.
We explore the following representations:

\textbf{Raw Representations} are the raw actions and rewards observed during training. 
These representations are constructed by simply concatenating the sequence of actions taken by the agent and the corresponding rewards obtained in each iteration.

\textbf{Catch22 Representation} are time-series features extracted from the raw actions and rewards observed during training.
These features capture a broad spectrum of time-series characteristics, including the distribution of values in the time series, linear and nonlinear temporal autocorrelation properties, scaling of fluctuations, and other relevant properties.
Another advantage to using time-series features is the ability to characterize and compare variable-length episodes.
We use the catch22~\cite{catch22} library to extract 22 time-series features from the observed sequences of actions and rewards together with mean and standard deviation, resulting in 24 features.
Note that we could use any other time-series~features.

Both representations (raw and catch22) can also be combined with instance features describing the problem instance and are not directly related to the behavior of the RL agent.
An example of such features can be the slope and shift of a sigmoid problem instance.

\paragraph{SELECTOR}
We execute the SELECTOR methodology using the different aforementioned representations to represent the instances from the training set.
We use the Dominating Sets (DS; ~\cite{dominating_set}) and Maximal Independent Set (MIS; ~\cite{maximal_independent_set}) algorithms with different similarity thresholds, specifically, $0.7, 0.8, 0.9$, and $0.95$.

\section{Experiments}
\label{sec:experiments}
For evaluating our method, we rely on the benchmark library DACBench~\cite{eimer-ijcai21a}, which features \ac{DAC} benchmarks from different AI domains.
We first cover the evaluation protocol, then the \ac{DAC} benchmarks used, Sigmoid and CMA-ES, and finally, detail the training of the \ac{RL} agent.

\paragraph{Evaluation Protocol}
Our overall objective is to assess the generalization performance on the test instance set $\testset$.
Therefore, we evaluate the agent trained on the full, original train instance set $\trainset$ and the agent trained on the subselected set $\selset \subseteq \trainset$ once again on the test instance set $\testset$.
For an empirical upper limit to performance on the test instance set, we additionally train \emph{Instance-Specific Agents (ISAs)}.
Each ISA is an \ac{RL} agent trained on one instance of the test instance set and evaluated on that specific instance, serving as a reference.
This construction of ISA exploits the notorious property of the RL agents to overfit their training instance:
\emph{Each ISA demonstrates the possible reward that an RL agent can accumulate when trained solely on this instance}.
In other words, they serve as an empirical performance upper bound that should be hard to achieve for a DAC agent being trained across a variety of training instances.
In addition, we also compare to \ac{RL} agents trained on $5$ random subsets of $10\%$ of the train instance set $\trainset$, which is a similar fraction of instances selected by SELECTOR.

We perform experiments on the Sigmoid benchmark, where a Sigmoid curve with varying slope and shift should be approximated, and on CMA-ES, where the step-size $\sigma$ is adapted.
In the following paragraphs, we further explain these benchmarks.

    

\paragraph{Sigmoid}
This benchmark challenges \ac{DAC} agents to approximate a Sigmoid function in different dimensions.
It is an artificial white-box benchmark that was proposed to study DAC with full control over the application~\cite{biedenkapp-ecai20a}.
A Sigmoid function is characterized by its shift and slope and has function values between $0$ and $1$.
Actions are discrete and evenly space the interval $[0,1]$.
For example, for an action space of $5$ actions, the actions would be $a \in \{0, 0.25., 0.5, 0.75, 1\}$.
We approximate Sigmoids in two dimensions, with $5$ and $10$ actions, respectively.
The state features consists of the remaining budget, the shift and slope for each dimension, and the action for each dimension.
The difficulty of the problem can be increased by increasing the dimensionality.
The training and test instance sets comprise $300$ instances of two-dimensional~Sigmoids.



\paragraph{CMA-ES}
CMA-ES (Covariance Matrix Adaption Evolution Strategy)~\citep{hansen-eda06a} is an evolutionary algorithm for continuous black-box problems which can be non-linear and non-convex.
In DACBench~\cite{eimer-ijcai21a}, the step-size $\sigma \in [0,10]$ of CMA-ES can be adapted, which is a continuous action space.
Others adapt the step-size via a heuristic~\cite{igel-evoc07,hansen-corr08} or guided policy search~\cite{shala-ppsn20a}.
As a state, the RL agent receives the generation size, the current step-size $\sigma$, the remaining optimization budget, as well as the function and instance ID.
The reward is the negative minimum function value observed so far since CMA-ES is a minimizer and the RL agent is a maximization algorithm.
The train and test instance set comprises ten synthetic blackbox optimization benchmarking (BBOB) functions~\cite{hansen-oms20a} -- Sphere, Ellipsoidal, Rastrigin, Büche-Rastrigin, Linear Slope, Attractive Sector, Step Ellipsoidal, original and rotated Rosenbrock and Ellipsoidal.
All of these functions are either separable or have low or moderate conditioning, except for the last one with high conditioning, $\in \mathbb{R}^{10}$.
The train set features four instances of each function, and the test set one instance.

\paragraph{Training Details}
We repeat our training and evaluation pipeline for $10$ random seeds.
Our training details are as follows: 
We train a PPO~\cite{schulman-arxiv17a} agent for $10\,000$ environment steps in Sigmoid, equaling $1\,000$ episodes, with each episode having a length of $10$. 
For CMA-ES, we train the agent for $1\,000\,000$ steps. 
However, here, we have variable episode lengths.
We evaluate each trained agent with $10$ evaluation episodes per instance.
Based on the evaluation rollout data, we run SELECTOR $5$ times and normalize the agent's performance per instance.
We then compute bootstrapped mean, median, and IQM with $5\,000$ samples using the library rliable~\cite{argawal-neurips21a} for the evaluation performance.
We additionally use fANOVA~\cite{hutter-icml14a} with standard settings to analyze the sensitivity of SELECTOR to its own hyperparameters, namely feature types, the method of selection, the source of features, and the threshold.

\paragraph{Instance representation and selection}
The chosen benchmark suites encompass training RL in distinct environments: one involving discrete actions (Sigmoid) and the other involving continuous actions (CMA-ES).
We employ different representations to depict the behavior of the RL agent.
Based on the actions (A) and rewards~(R) recorded on evaluation rollouts, we either use the raw (flattened) vectors for fixed-length episodes or the catch22 time-series features for variable-length episodes.
We can also concatenate action and reward vectors (RA) and add instance features~(I) if applicable.

In both benchmark suites and their respective instance representations, we employ the SELECTOR method (both MIS and DS with similarity thresholds $\in \{0.7, 0.8, 0.9, 0.95\}$ for creating the graph) to choose subsets of instances for retraining the RL agent.

\subsection{Results and Discussion}
\label{sec:results}

\begin{figure}[t]
    \centering
    \includegraphics[width=\linewidth]{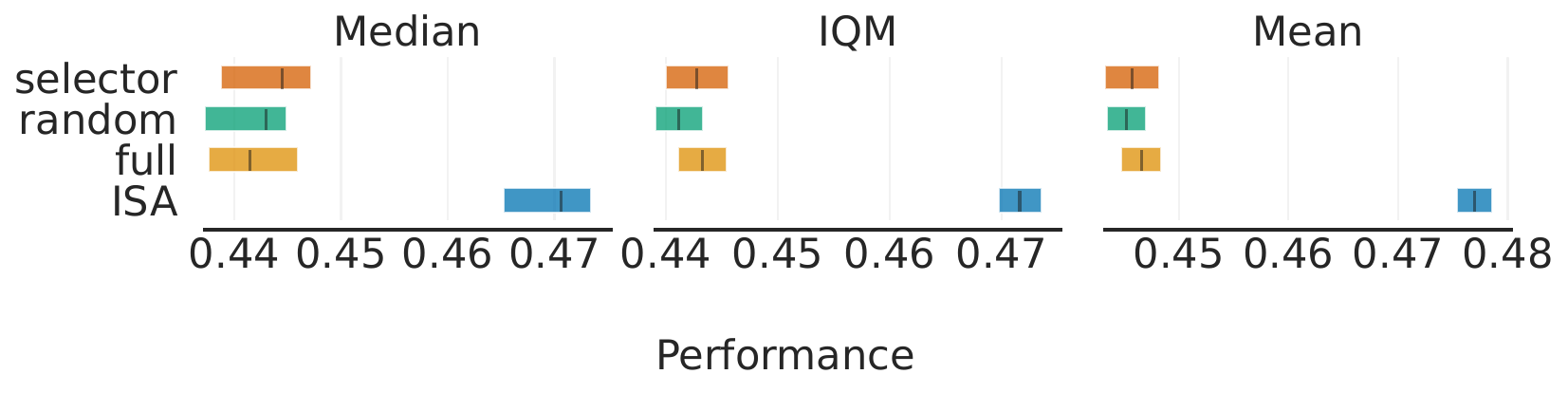}
    \caption{Sigmoid Performance}
    \label{fig:sigmoid_performance}
\end{figure}

\begin{figure}[t]
    \centering
    \includegraphics[width=\linewidth]{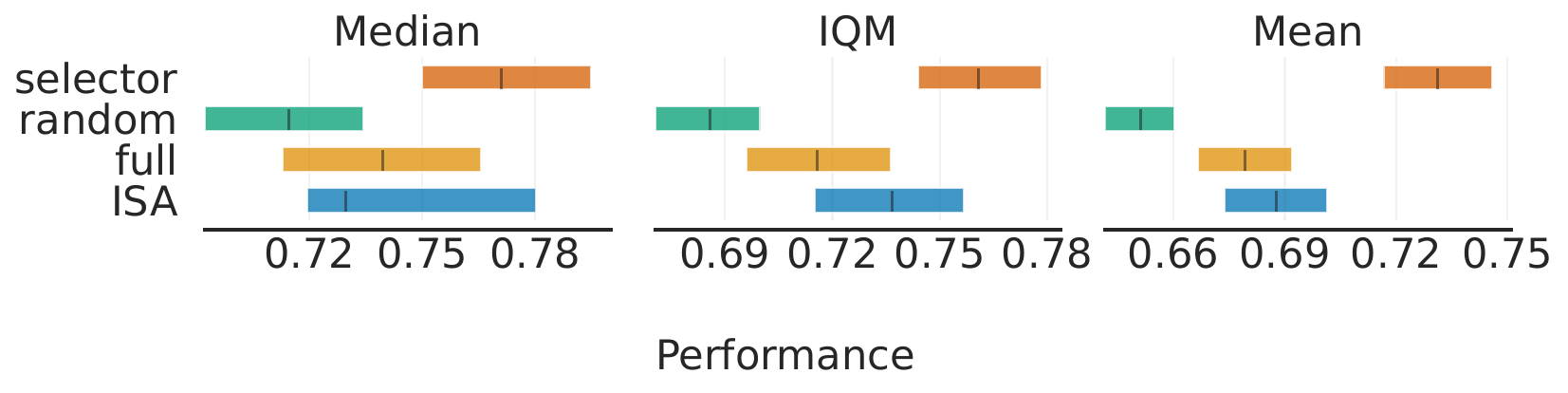}
    \caption{CMA-ES Performance}
    \label{fig:cmaes_performance}
\end{figure}

On both benchmarks, Sigmoid and CMA-ES, training on subselected instances from SELECTOR generalizes better to the test instance set than training on the full instance set, see~
 \cref{fig:sigmoid_performance} for Sigmoid and \cref{fig:cmaes_performance} for CMA-ES.
First of all, this supports our hypothesis that training an DAC agent with RL on a well-constructed subset of instances can be better than simply training on an arbitrary instance set. 
Secondly, the extraction of trajectory information is sufficiently informative to construct this set. 

Interestingly, the performance of the ISA for CMA-ES is worse than the performance of SELECTOR.
Initially, our aim was to construct ISA so that we get an empirical approximation of a theoretical upper limit; thus our DAC agent on SELECTOR should not be able to outperform ISA.
We hypothesize that the diversity in trajectories from multiple instances instead of only one instance allows the optimization process of the \ac{RL} agent to escape potential local minima in policy space that the ISA agents get stuck in.
This corroborates the successful methodology of learning the step size with guided policy search~\cite{shala-ppsn20a}, where they guide the optimization and start from a suitable point in the policy space.


Depending on the benchmark we observe different best performing variants of SELECTOR.
According to the IQM, SELECTOR with (MIS, Catch22, R, $0.7$) for Sigmoid and SELECTOR with (DS, Catch22, R or RA, $0.8$) for CMA-ES performed best.
So, it is important to study the hyperparameter (HP) sensitivity of our approach. 
For Sigmoid the type of representation is important, using only actions or combinations with reward yields best results.
The other HPs do not have a major impact on Sigmoid.
For CMA-ES, the subselection method on the similarity graph (DS or MIS) is the most important HP.
Again, representations using actions and rewards together works best.
A reasonable robust and general choice would be to use rewards and actions as features sources combined with~DS.

\begin{figure}[h]
    \centering
    \includegraphics[width=0.44\linewidth]{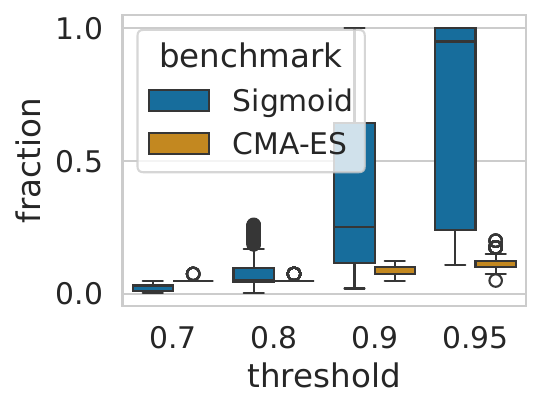}
    \includegraphics[width=0.55\linewidth]{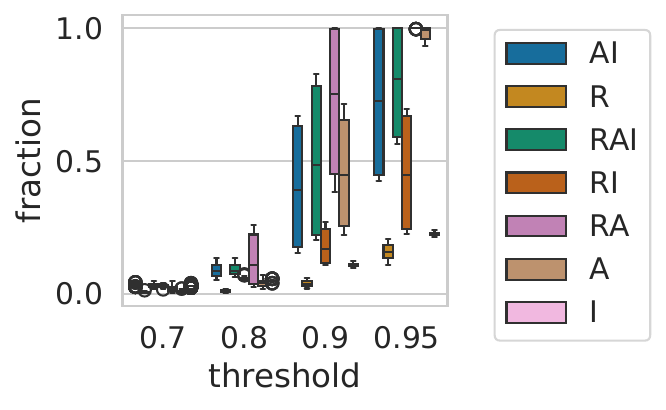}
    \caption{(Left) Size of subselected instance set for different SELECTOR thresholds per benchmark. (Right) Size of subselected instance sets for Sigmoid for different representations. }
    \label{fig:thresholds}
\end{figure}

The size of the instance set shows strong variation for the threshold of SELECTOR for Sigmoid, but not so much for CMA-ES, as shown in \cref{fig:thresholds}.
Peaking closer into Sigmoid, \cref{fig:thresholds} (right) indicates that instance features and trajectory features are not very correlated.
A small instance set with a high threshold induces a dense graph, i.e. instances are pretty similar in terms of instance features which does not necessarily mean trajectory features are~similar.

\begin{figure}[h]
    \centering
    \includegraphics[width=0.7\linewidth]{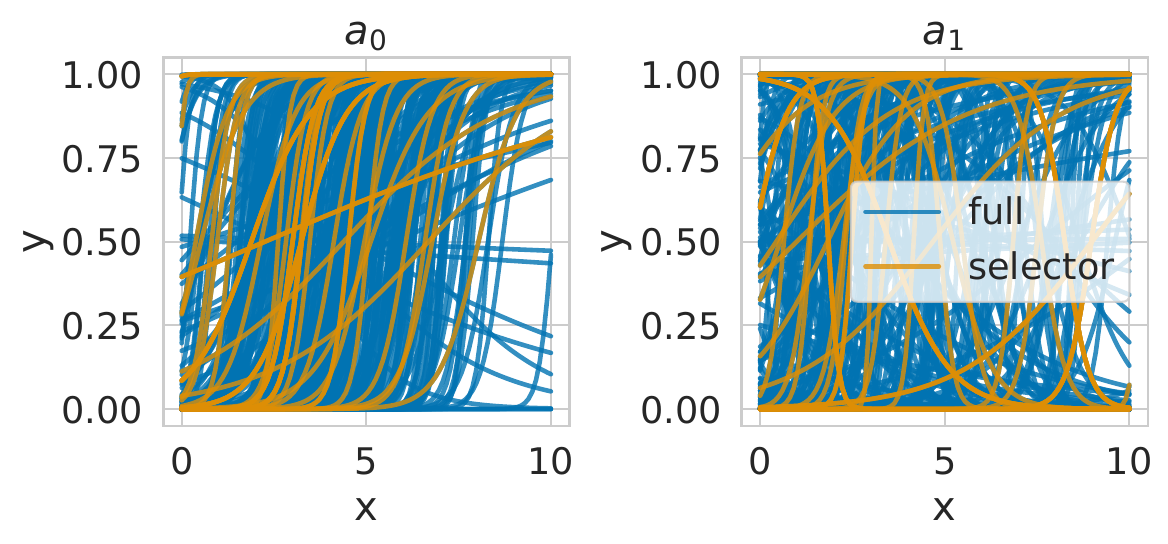}
    \caption{Selected instances by SELECTOR for Sigmoid. A small but diverse set of instances is selected.}
    \label{fig:sigmoid_selected_instances}
\end{figure}

In addition, the instances selected by SELECTOR evenly cover the full instance set, capturing the diversity that is most apparent for the second dimension (\cref{fig:sigmoid_selected_instances}).
For CMA-ES, often only one instance of the BBOB functions 7, 8, 9 is selected.
These functions have a more complex local structure compared to the first functions but still are similar in global structure, rendering them suitable to represent the instance set.

\paragraph{Limitations and Future Work}
One limitation of our method is that it requires training the \ac{RL} agent twice as well as training SELECTOR.
We plan to investigate the benefits that can be potentially gained from \emph{early-stopping}, such as only training the agent for half of the training budget. 
Potentially, the benefits of SELECTOR could be attained in the same training budget as a standard baseline agent.
We additionally plan to meta-learn well-performing presets for SELECTOR to create a truly end-to-end training and selection pipeline.
Lastly, we would like to approach handling instances also at the level of the \ac{RL} algorithm:
For benchmarks like CMA-ES, we have problems with different reward scales, potentially hindering learning, which we could normalize per instance.

\section{Conclusion}

In this work, we demonstrate the potential of instance selection in enhancing the generalization capabilities of \acf{RL} for \acf{DAC}. 
We first train an \ac{RL} agent on a train set of instances and then generate rollout trajectories by evaluating the trained agent on the same set of instances.
Since these trajectories capture the agent's behavior on the training instances, we use this data to create time-series features that capture the \emph{dynamic} behavior of the \ac{RL} policy.
We then subselect a representative set of training instances and retrain the \ac{RL} agent on these instances to obtain better generalization performance on unseen new instances.
By meticulously selecting representative instances for training, we not only address the challenge of overrepresentation in training instances but also demonstrate superior performance to agents trained on specific instances on CMA-ES. 
Our approach marks a step forward in the application of \ac{RL} to \ac{DAC}, offering a scalable solution that can adapt to the ever-changing complexities of hyperparameter control using \ac{RL}. 

\section{Acknowledgements}
Funding in direct support of this work: Slovenian Research Agency: research  core  funding  No. P2-0098, young researcher grants No. PR-12393 to GC and No. PR-12897 to AN, project No. J2-4460, and a bilateral project between Slovenia and Germany grant No. BI-DE/23-24-003.  DAAD: 57654659.

\bibliographystyle{ACM-Reference-Format}
\bibliography{bib/strings,bib/lib,bib/local,bib/proc}



\end{document}